%% file: root.tex
\documentclass[letterpaper, 10 pt, conference]{ieeeconf}

\makeatletter
\def\endthebibliography{%
  \def\@noitemerr{\@latex@warning{Empty `thebibliography' environment}}%
  \endlist
}
\makeatother
 
\IEEEoverridecommandlockouts
\usepackage{cite}
\usepackage{amsmath,amssymb,amsfonts}
\usepackage{algorithmic}
\usepackage{graphicx}
\usepackage{textcomp}
\usepackage{xcolor}
\usepackage{todonotes}
\usepackage{amsmath}
\usepackage{leftidx}
\usepackage{siunitx}
\usepackage{multirow}
\usepackage{footnote}
\usepackage[utf8]{inputenc}
\usepackage{array}
\usepackage{mdwmath}
\usepackage{mdwtab}
\usepackage{eqparbox}
\usepackage{url}
\usepackage{cite}
\usepackage{amsmath,amssymb,amsfonts}
\usepackage{algorithmic}
\usepackage{textcomp}
\usepackage{lettrine}
\usepackage{caption}
\usepackage{subcaption}
\usepackage[export]{adjustbox}
\usepackage{amsmath}
\usepackage{leftidx}
\usepackage{siunitx}
\usepackage{multirow}
\usepackage{balance}
\usepackage[export]{adjustbox}
\usepackage{booktabs}
\usepackage{bbold}
\usepackage{hyperref}
\newcommand{\rulesep}{\unskip\ \vrule\ }
\def\BibTeX{{\rm B\kern-.05em{\sc i\kern-.025em b}\kern-.08em
    T\kern-.1667em\lower.7ex\hbox{E}\kern-.125emX}}
\begin{document}

\title{\LARGE \bf Robot Localization using Situational Graphs \\ and Building Architectural Plans}

\author{Muhammad Shaheer, Hriday Bavle, Jose Luis Sanchez-Lopez, and Holger Voos 
\thanks{*This work was partially funded by the Fonds National de la Recherche of Luxembourg (FNR), under the projects C19/IS/13713801/5G-Sky, by a partnership between the Interdisciplinary Center for Security Reliability and Trust (SnT) of the University of Luxembourg and Stugalux Construction S.A.}
\thanks{Authors are with the Automation and Robotics Research Group, Interdisciplinary Centre for Security, Reliability and Trust, University of Luxembourg. Holger Voos is also associated with the Faculty of Science, Technology and Medicine, University of Luxembourg, Luxembourg.\tt{\small{\{muhammad.shaheer, hriday.bavle, joseluis.sanchezlopez, holger.voos\}}@uni.lu}}%
}

\maketitle
\input{abstract}

\input{introduction}

\input{related_works}
\input{proposed_method}
\input{experimental_evaluation}
\input{conclusion}
\bibliographystyle{IEEEtran}
\bibliography{Bibliography}

\end{document}

%% file: abstract.tex
\begin{abstract}

\label{abstract}
 Robots in the construction industry can reduce costs through constant monitoring of the work progress, using high precision data capturing. Accurate data capturing requires precise localization of the mobile robot within the environment. In this paper we present our novel work on robot localization which extracts geometric, semantic as well as the topological information from the architectural plans in the form of walls and rooms, and creates the topological and metric-semantic layer of the Situational Graphs (\textit{S-Graphs}) \cite{s_graphs} before navigating in the environment. When the robot navigates in the construction environment, it uses the robot odometry and the sensorial observations in the form of planar walls extracted from the 3D lidar measurements, to estimate its pose relying on a particle filter method, by exploiting the previously built situational graph and its available geometric, semantic and topological information. We validate our approach in both simulated and real datasets captured on actual on-going construction sites presenting state-of-the-art results when comparing it against traditional geometry based localization techniques.
\end{abstract}

%% file: introduction.tex
\section{Introduction}
\label{introduction}

 Robots are starting to be used on construction sites to perform various tasks such as autonomous data capturing to continuously monitor the evolution of the construction site. While construction environments are challenging because of their dynamic nature, as they can change significantly over a span of few days, they have some features that are common in most of them, for example walls, rooms, or pillars. Moreover, many companies are now integrating digital tools such as Building Information Modelling (BIM) \cite{bim} in their project design and execution phases. BIM contains geometric, semantic, topological and physical information about each and every element of the building. Such crucial information is invaluable for a robot, which can use this knowledge for localization and understanding the construction site around it. Nevertheless, nowadays the robots are unable to exploit the power of these digital plans. When they navigate in the environment, they either build their internal maps from scratch, which makes it difficult for them to know where they are, or they require fiducial markers to localize properly. 
\begin{figure}[!ht]
    \centering
    \includegraphics[width=0.5\textwidth]{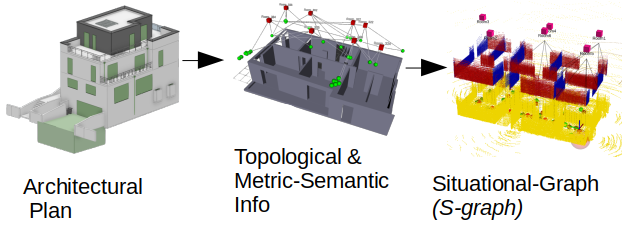}
    \caption{Visualization of the proposed approach. First, the topological, semantic and geometric information is extracted from an actual architectural plan of a building, made in Revit. A topological layer and a metric-semantic layer of \textit{S-Graph} is created from this information. This information along with the initial pose estimate from a particle filter are used to localize the robot through generated \textit{S-Graph} for the environment.}
    \label{fig:scene_graph}
\end{figure}

 Most of the on-going construction sites can be well represented using walls and rooms, to provide the robot with pre-existing knowledge of the environment. Robots utilizing only geometric information for localization within a construction site may not be able to localize correctly due to the fact the the on-going site does not resemble yet the architectural plans. Semantic information can help improve the localization, but still can suffer inaccuracies due to large deviation of the robot pose estimate from its actual estimate and similar semantic information in case of repetitive environments. Thus, research is still required to combine the complete geometric, semantic and topological information provided by the BIM models, to robustly localize a robot within an on-going construction environment.

This paper proposes a novel method in this direction to localize a robot on the construction environments using information extracted from architectural plans, and Situational Graphs (\textit{S-Graphs}) \cite{s_graphs} \footnote{\url{https://www.youtube.com/watch?v=8aw-b2PWRwQ}}. We extract this geometric (planar walls), semantic (type of planar walls) and topological information (room constraining four walls) from the architectural plans, which is first utilised to estimate the initial pose of the robot in the building, using Monte Carlo Localization. Once the initial pose is estimated, the robot uses the knowledge from topological and metric-semantic layers of  \textit{S-Graph}, created in the first step, to further localize itself as it navigates in the environment.
Our main contributions in this paper are: 
\begin{itemize}
    \item Extracting geometric, semantic and topological information of a building from its existing architectural plans useful to augment the robot's pre-existing knowledge of the environment.
    \item Particle filter based estimate of the global pose of the robot within the architectural plans using the extracted geometric, semantic as well as topological information.
    \item Hierarchical \textit{S-Graphs} \cite{s_graphs} based localization of the robot, utilizing the global estimate from the particle filter and the extracted information from the architectural plans.
\end{itemize}

\begin{figure}[!ht]
    \centering
    \includegraphics[width=0.43\textwidth]{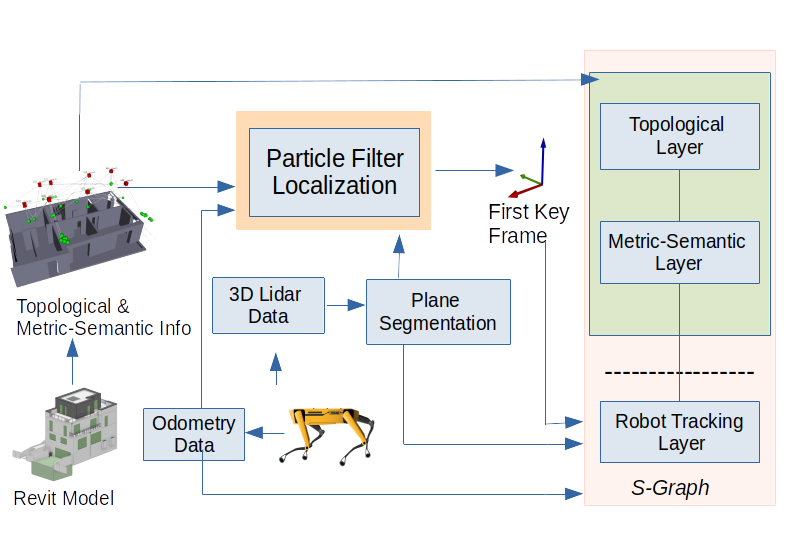}
    \caption{Pipeline of our proposed localization method, architectural plan of a building created in Revit is used to extract topological, semantic and geometrical information of the building. With this information the topological and metric-semantic layers of the \textit{S-Graph} are created which are used by particle filter to estimate the initial pose of the robot. After estimating the initial pose, \textit{S-Graph} starts creating the robot-tracking layer and connecting it with the top two layers optimizing the entire \textit{S-Graph}.}
    \label{fig:system_architecture}
\end{figure}

%% file: related_works.tex
\section{Related Works}
\label{related_works}

\subsection{Architectural Plans based Localization}

In robotics use of 2D CAD based models is very common to generate a 2D map of the environment and to localize a robot using the on-board range sensors \cite{radar_localization_cad}, \cite{cad_based_localization}. \cite{robot_loc_in_floor_plans} utilizes a monocular system to localize the robot in the 2D floor plans extracting the room layout edges. \cite{loc_parking_lot}, \cite{loc_zimmerman} incorporate semantic textual cues into the 2D floor plans to further improve the localization accuracy. 

BIM on the other hand provides additional semantic and topological information as compared to 2D CAD plans which can be exploited to improve the planning and localization of mobile robots.  
Authors in \cite{Siemitkowska2013BIMBI}, \cite{bim_path_plannin_1}, \cite{bim_path_planning_2} utilize BIM information for extracting 3D maps utilized by path-planner but do not include this information for robot localization. Authors in  \cite{localization_bim_with_images} only present a small case study localizing the robot through image matching between real images and images from BIM plans, but only restricted to a certain perspective. \cite{bim_localization} present an approach extracting information from BIM, splitting it into different floor levels and then converting to a relevant format used by a state-of-the-art pose-graph SLAM algorithm. Since the pose-graph algorithm cannot take as input an entire 3D mesh of the area, the authors have to create an additional module to generate trajectories within the mesh storing pose and sensor data at different intervals to later input it the SLAM for localization. Whereas in \cite{bim_localization_case_study}, authors present a case study converting BIM into 3D meshes at different floor levels and using a simple Iterative Closest Point (ICP) algorithm to localize the robot, thus depending highly on the metric ICP algorithm for convergence, which can show inaccuracies if an on-going construction site does not yet resemble the BIM plans. Compared to previous approaches, \cite{connecting_semantic_bim} extract semantic information in addition to geometric information from the BIM plans to localize the robot using an on-board 2D LiDAR and pose-graph optimization, but the authors do not consider additional topological information connecting different walls to a room etc.    

\subsection{Scene Graph based Localization}

Scene graphs is an emerging research to represent the scene in several geometric, semantic and topological dimensions \cite{3d_scene_graph}, \cite{scene_graph_fusion}, \cite{dynamic_scene_graph}.
\textit{S-Graphs} \cite{s_graphs} is a real-time optimizable hierarchical scene graph which represents the environment in geometric, semantic and topological dimensions. \textit{S-Graphs} is capable of not only representing the environment as a scene graph but also simultaneously optimize all the elements within it including the robot pose. All of these methods need the robot to navigate the environment in order to create thee scene-graphs, which means they need to know the starting point of the robot beforehand. They lack the capability of performing global localization. Some methods have also emerged exploiting scene graphs for localization such as \cite{semantic_loop_closure}, \cite{x_graph}, \cite{semantic_histogram} performing graph based localization utilizing spatial information between the neighbouring landmarks. Similarly authors in \cite{scene_graph_loc} create a global topological semantic map of the environment and later use similarity between semantic scene graphs to localize the robot. \cite{global_loc_gnn} combine traditional 3D object alignment with graph based matching to perform global localization in small sized areas such as rooms. 

None of these approaches use the knowledge of the environment available in the form of digital plans. Therefore, inspired from the current state-of-the-art, we present our novel approach which combines BIM and scene graphs to provide robust localization of robots. Our approach  extracts information from BIM, but we can efficiently input the extracted geometric, semantic and topological data to a hierarchical graph \cite{s_graphs} without the need of moving the robot or generating additional trajectories within the virtual 3D mesh to create the map of the environment. 

%% file: proposed_method.tex
\section{Proposed Approach}
\label{proposed_approach}
\subsection{Overview}
Our approach can be divided into three main modules: building information extraction module, particle filter based localization module, and \textit{S-Graphs} \cite{s_graphs} localization module. 
We define the global world frame of reference $W$ which coincides the origin frame of reference defined within the architectural plans. The odometry of the robot frame $R_t$ is computed with respect to the odometry frame $O$. Lidar measurements from the 3D LiDAR are received in the LiDAR frame $L_t$. The building information extraction module extracts all the relevant information from the architectural plans. The particle filter utilizes the geometric/semantic information i.e mapped planar walls along with the type of wall (vertical $x$/$y$-axis) as well as the topological information of the rooms, to estimate the initial alignment between the odometry frame and the world frame $\leftidx{^W}{\mathbf{x}}_{t_{1}O}$. The \textit{S-Graphs} localization module receives the mapped walls and the room information to generate beforehand the top two layers of the \textit{S-Graphs} (i.e metric-semantic layer and topological layer), which start getting connected to the first layer (robot tracking layer) which is generated after the reception of the initial pose $\leftidx{^W}{\mathbf{x}}_{t_{1}O}$ and at different intervals as the robot navigates. An overview of our approach is shown in Fig.~\ref{fig:system_architecture}.  

\subsection{Building Information Extraction} \label{sec:bim}
Architecture plans of  buildings are made in various formats. In this paper we use the International Foundation Class (IFC) format \cite{ifc}. An IFC file is a Building Information Model (BIM) file which is platform independent, meaning that an IFC file can be used in any BIM software. 
We propose to extract geometrical, semantic and topological information from the IFC file and exploit it for localization and environmental understanding by the robot. 
\subsubsection{Floor Information Extraction} \label{BIM}
A building can have multiple floors and therefore, to localize a robot accurately in the building using BIM information, accurate floor/storey of the building needs be determined. The information about the floors of the building is defined in an IFC file by a parameter called \textit{IfcBuildingStorey}. This parameter further has several attributes but the most important one in \textit{Elevation} attribute. \textit{Elevation} attribute, as the name suggests, gives the elevation of the floor from the ground. Therefore, if the robot is capable of sensing its height from the ground and we have already extracted the heights of all floors of the building, it is straight forward to determine the exact floor on which the robot is. However, at this stage we are not performing multi-floor mapping of the environment.

\subsubsection{Rooms Information Extraction}
Once we determine the floor of the building, next step is to determine the exact location of the robot on that floor. In IFC files, the rooms are defined by a parameter named \textit{IfcSpace}. This parameter further has more attributes which give the information about the area and the position of the room $[\leftidx{^W}{\rho_x}, \leftidx{^W}{\rho_y}]$ in world frame. 

\subsubsection{Walls Information Extraction}
Walls are arguably the most common elements to be found in an IFC file. Each wall has its own geometrical and physical properties. Each of these walls can be used as landmarks as it has a fixed defined position, and thickness. In IFC file, the wall is defined by \textit{IfcWall} and \textit{IfcWallStandardCase}. We extract the location, orientation and thickness of each wall from the IFC file.
We use the general form of a plane equation to extract the perpendicular distance of the wall $\leftidx{^{W}}{\mathbf{\pi}}$ in the world frame $W$ as: 

\begin{equation} \label{eq:general_plane_eq}
   \leftidx{^{W}}d =  -1 \cdot (a \cdot \leftidx{^{W}}n_{x} + b \cdot \leftidx{^{W}}n_{y} + c \cdot \leftidx{^{W}}n_{z})
\end{equation} 

\noindent Here $a$, $b$, and $c$ are the $x$, $y$, and $z$ coordinates of the starting point of the wall respectively, and \mbox{$\leftidx{^{W}}{\mathbf{n}_{\pi}} =$ [$\leftidx{^{W}} n_x, \leftidx{^{W}}n_y, \leftidx{^{W}}n_z$]} are the normal orientations of the wall. 
In the IFC file each wall is considered as a single wall, but for the robot they are two different walls with opposite orientation known as the double sided issue \cite{lidar_slam_with_plane_adj}. We solve this issue by creating a duplicate wall for $\leftidx{^{W}}{\mathbf{\pi}}$ with opposite orientation and increasing its perpendicular distance $\leftidx{^{W}}d$ by the wall thickness as:

\begin{equation} \label{eq:opposite_wall}
    \leftidx{^{W}}{\mathbf{\pi}^\mathcal{D}} =  \begin{bmatrix}
     -\leftidx{^{W}}{\mathbf{n}_{\pi}} \\ 
     \leftidx{^{W}}{{d}_{\pi} + \mathcal{T}_{\pi}} 
    \end{bmatrix}
\end{equation}

Here  $\leftidx{^{W}}{\mathbf{\pi}^\mathcal{D}}$ is the opposite wall of  $\leftidx{^{W}}{\mathbf{\pi}}$ and $\mathcal{T}_{\pi}$ is the thickness of wall $\leftidx{^{W}}{\mathbf{\pi}}$

\subsubsection{Topological and Metric-Semantic Layer Generation} \label{subsec:scene_graph}
We create the topological and metric-semantic layers of \textit{S-Graph} of a building by using this extracted information from BIM model. The topological layers consists of \textit{Rooms} of the building, whereas the metric-semantic layers consists of the \textit{Walls}. These two layers of the \textit{S-Graph} are then used by the robot to extract information and localize itself in the environment, further explained in Section.~\ref{sec:s_graphs_localization}

\subsection{Localization using Particle Filter}
\label{particle_filter}
We introduce a novel topological information factor in particle filter based localization. The particles are matched with only the walls of the rooms in which they lie in, instead of matching them with all the walls of that storey of the building. The particles are uniformly distributed within the boundaries of the environment, in world reference frame $W$. The the state of each particle is defined as $\boldsymbol{x}^{n} = [x, \ y, \ z, \ \textbf{q} ]^T$
where $n$ is from 1 to $N$, and $x$, $y$, $z$ are positions of a particle in in world frame $W$, and, $\textbf{q} = [q_{x}, \ q_{y}, \ q_{z}, \ q_{w}]$ is a quaternion representing the orientation of the particle in $W$ frame.

\subsubsection{Observations} In our case observations are detected and segmented planar walls from the 3D LiDAR measurements as the robot navigates within its environment. The plane extraction algorithm is inspired by \cite{s_graphs} and uses a sequencial RANSAC to extract all the planar coefficients of all the planar surfaces at a given time instant. The planes are detected in LiDAR frame $L_t$ and converted to their Closest Point (CP) representation $\leftidx{^{L_t}}{\mathbf{\Pi}}$ \cite{lips} are defined as: 

\begin{equation}
    \leftidx{^{L_t}}{\mathbf{\Pi}} = \leftidx{^{L_t}}{\mathbf{n}}^{\prime} \cdot \leftidx{^{L_t}}{d}^{\prime} 
\longrightarrow
    \begin{bmatrix}
     \leftidx{^{L_t}}{\mathbf{n}} \\ 
     \leftidx{^{L_t}}{{d}} 
    \end{bmatrix} =  \begin{bmatrix}
     \leftidx{^{L_t}}{\mathbf{\Pi}} / \| \leftidx{^{L_t}}{\mathbf{\Pi}} \| \\ 
     \| \leftidx{^{L_t}}{\mathbf{\Pi}} \| 
    \end{bmatrix}
\end{equation}

\noindent where $\leftidx{^{L_t}}{\mathbf{n}}=[\leftidx{^{L_t}}n_x, \ \leftidx{^{L_t}}n_y, \ \leftidx{^{L_t}}n_z]^\top$ is the plane normal and $\leftidx{^{L_t}}{d}$ is the distance to the origin, both in the LiDAR frame.

\subsubsection{Landmarks} 
We use walls of the building, extracted from the architectural plans, as landmarks in our approach. These walls are represented as planar surfaces in the $W$ frame. The planar coefficients of all the walls are obtained using Eq.~\ref{eq:general_plane_eq} and Eq.~\ref{eq:opposite_wall}. 



\subsubsection{Prediction}
In the prediction step, the particle pose is updated using 3D robot pose obtained from the odometry either provided by the encoders of robot platform or using the measurements of the 3D LiDAR. The state of each particle is propagated from time $t-1$ to $t$ based on a motion model defined as:
\begin{equation}
   \mathbf{x}_{t}^{[n]} \sim p(\mathbf{x}_{t} \ | \ \mathbf{x}_{t-1}^{[n]}, \mathbf{u}_{R_{t}})
\end{equation}

\noindent $\mathbf{x}_{t-1}^{[n]}$ is the pose of $n$-th particle and $\mathbf{u}_{R_{t}}$ is the incremental pose of robot between time instance $t-1$ and $t$:
\begin{equation}
    \mathbf{u}_{R_{t}} =  \leftidx{^O}{\mathbf{x}}^{-1}_{R_{t}} \oplus  \leftidx{^O}{\mathbf{x}}_{R_{t-1}}
\end{equation}

\noindent $\leftidx{^O}{\mathbf{x}}_{R_{t}}$ and $\leftidx{^O}{\mathbf{x}}_{R_{t-1}}$ are the robot poses at time $t$ and $t-1$ in odometry frame $O$ respectively.

\subsubsection{Data Association} 
Our novel data association step, exploits not only the geometric/semantic information but also the topological information, exploiting the available top layers of the S-Graph and performing the data association in two different steps, as follows:

\textbf{Room Association.} The BIM model gives us the starting points $[\leftidx{^W}{\rho_x}, \leftidx{^W}{\rho_y}]$  of the rooms in $W$ frame, and the areas of the rooms, as well as the walls constrained by each room, which can help define the boundaries of each room. We thus  check the position of each particle, and, if it lies within the boundaries of a room, the currently observed planes are matched only to the walls belonging to that particular room instead of matching that particle's observations with all the landmarks of the building. This procedure significantly reduces the ambiguity in data association, computation time and cost as well.

\begin{figure*}[!ht]
\centering
\begin{subfigure}{0.28\textwidth}
\centering
\includegraphics[width=0.45\textwidth]{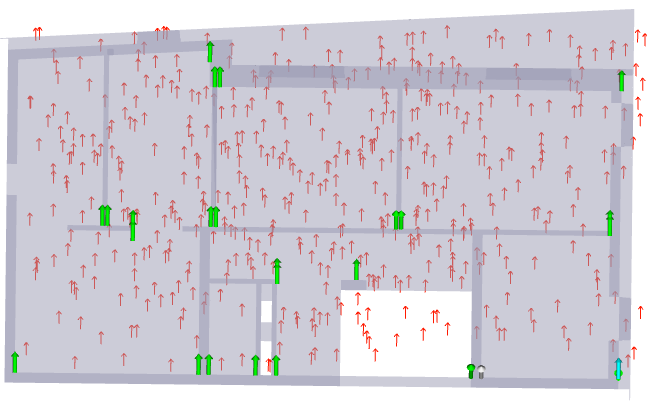}
\caption{Initialization}
\label{fig:pf}
\end{subfigure}
\rulesep
\begin{subfigure}{0.28\textwidth}
\centering
\includegraphics[width=0.45\textwidth]{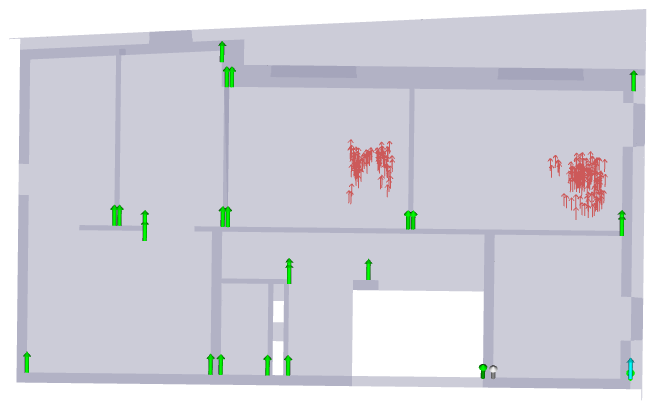}
\caption{Clustering}
\end{subfigure}
\rulesep
\begin{subfigure}{0.28\textwidth}
\centering
\includegraphics[width=0.45\textwidth]{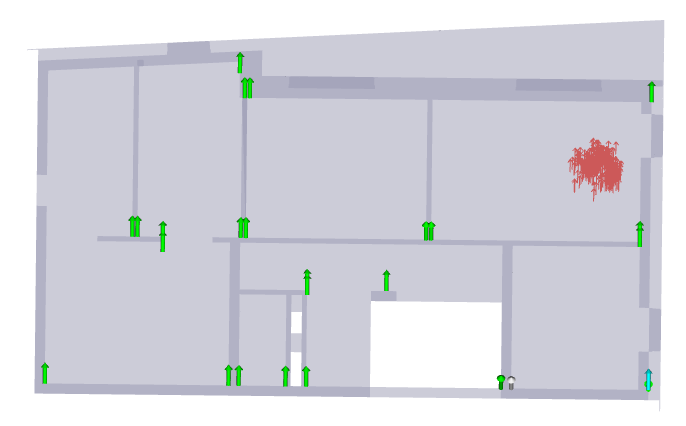}
\caption{Convergence}
\end{subfigure}
\label{fig:}
\caption{Top view of the particle filter localization with topological rooms information  (a) The particles are initialized in entire floor. (b) Particles form two clusters after update step. Note the 'clusters' are formed in 2 rooms with similar geometry. (c) The particles successfully converge in the correct room and the initial pose is published.}
\label{fig:paricle_filter_steps}
\end{figure*}

\textbf{Wall Association.}
After the room association step, we perform wall association by matching the observations to the mapped planar walls of the associated room's. All the planar walls are semantically categorized either $x$ vertical wall (with normal coefficient $n_x$ greater) and $y$ vertical wall (with normal coefficient $n_y$ greater). As all the landmarks are defined in world frame $W$, for each $i$th particle with pose $\leftidx{^W}{\mathbf{x}}_{P_{t_i}}$, we transform the observed planes in LiDAR frame $L_{t}$ to the world frame $W$ by :

\begin{equation}
    \leftidx{^{W}}{\mathbf{\pi}^{\prime}} = \begin{bmatrix}
     \leftidx{^W}{\mathbf{n}^{\prime}} \\ 
     \leftidx{^W}{{d}^{\prime}} 
    \end{bmatrix} = 
    \leftidx{^W}{\mathbf{T}}_{\boldsymbol{\pi}_{t}}(\leftidx{^W}{\mathbf{x}}_{P_{t_i}})\begin{bmatrix}
     \leftidx{^{L_t}}{\mathbf{n}} \\ 
     \leftidx{^{L_t}}{{d}} 
    \end{bmatrix}
\end{equation}

Here $\leftidx{^W}{\mathbf{T}}_{\boldsymbol{\pi}_t}$ is the transformation matrix that transforms the normal and distance of detected planes in world frame $W$. In order to compute the matching error between the observed and the mapped landmarks of each category ($x$ and $y$ wall), we use the minimal plane parametrization \cite{s_graphs}, where each plane is represented as $\leftidx{^W}{\boldsymbol{\pi}} = [\leftidx{^W}\phi, \leftidx{^W}\theta, \leftidx{^W}d]$, $\leftidx{^W}\phi$ and $\leftidx{^W}\theta$ are the azimuth and elevation of the plane in world frame $W$. While matching, we check the error between observations and landmark as:
\begin{equation} \label{eq:plane_error}
    \boldsymbol{\varepsilon} = \| \leftidx{^{L_t}}{{\boldsymbol{\pi}}}_i - \leftidx{^{L_t}}{\tilde{\boldsymbol{\pi}}}_i \|^2_{\mathbf{\Lambda}_{\boldsymbol{\tilde{\pi}}_{i,t}}}
\end{equation}

If the error between an observed plane and a landmark wall is less than a threshold $ \boldsymbol{\epsilon} < th$, we associate the observation with that particular landmark, else the observation is rejected.  We use the Mahalanobis distance between each observed plane
and the landmark planes.

\subsubsection{Update}
Once the incoming observations are associated with the corresponding landmarks, the weight of the matching particle is calculated as:

\begin{equation}
     \omega_{i} = \mu \cdot \exp{(-(\phi_{diff} + \theta_{diff} + d_{diff}))}
\end{equation}

where,
\begin{equation}
    \phi_{diff} = \tau \cdot (\leftidx{^{W}}\phi^{\prime} - \leftidx{^{W}}\phi)
\end{equation}

and $\mu = {(2 \cdot \pi \cdot \sigma^2)}^{-1/2}$, $\tau = (2 \cdot \sigma^2)$. $\theta_{diff}$ and $d_{diff}$ are also defined similarly. The calculated weights are first normalized and then assigned to the respective particle. Re-sampling of the particles is then performed using importance sampling principle to draw $N$ samples with probability proportional to the particle's weight. 


\subsubsection{Initial Transformation Estimation}
Our implementation of the particle filter only estimates the initial transformation between the odometry frame $O$ and the world frame $W$, once all the particles converge. In order to verify if the particles have converged, we take the current pose of the particle with the highest weight at time $t$, and check difference from the average pose of all the particles. If the difference is below a defined threshold, we consider the particles converged. We then take the pose of the particle with the highest weight, and compose it with the inverse of the robot pose estimated by the odometry in frame $O$ at time $t$, to estimate the initial transformation between the odometry frame and the world frame:
\begin{equation}
    \leftidx{^W}{\mathbf{x}}_{t_{1}O} =  \leftidx{^W}{\mathbf{x}}_{P_{t}} \oplus \leftidx{^O}{\mathbf{x}}^{-1}_{R_{t}} 
\end{equation}

$\leftidx{^W}{\mathbf{x}}_{t_{1}O}$ is the initial transformation between frame $O$ and $W$, whereas $\leftidx{^O}{\mathbf{x}}_{R_{t}}$ is the robot pose in $O$ frame and $\leftidx{^W}{\mathbf{x}}_{P_{t}}$ is pose of the particle with highest weight at the time of convergence.

\subsection{S-Graph based Localization} \label{sec:s_graphs_localization}
 \textit{S-Graphs} detects planes, rooms and corridors as the robot navigates the environment to create the optimizable graph. 

\begin{itemize}
    \item First layer consists of the robot poses $\leftidx{^M}{\mathbf{x}}_{R_t}, \ t \in \{1, \hdots, T\}$ at $T$ selected keyframes.
    \item The second layer is the metric-semantic layer which consists of $P$ detected planes $\leftidx{^M}{\boldsymbol{\pi}}_{i}, \ i \in \{1, \hdots, P\}$.
    \item The third and final layer istopological layers which consists of  $S$ rooms $\leftidx{^M}{\boldsymbol{\rho}}_{j}, \ j \in \{1, \hdots, S\}$, and, $K$ corridors $\leftidx{^M}{\boldsymbol{\kappa}}_{k}, \ k \in \{1, \hdots, K\}$.
\end{itemize}  

In this work we create the top tow layers of \textit{S-Graphs} from the information extracted from BIM. We create the metric-semantic layer (planar wall nodes) and topological layer (room node with appropriate edges constraining the room with its planar walls) of the \textit{S-Graphs} for a given environment. 

When the particle filter (Section.~\ref{particle_filter}) provides the initial pose estimation, \textit{S-Graphs} starts creating its first robot-tracking layer and connects it with the metric-semantic and topological layers. 
\textit{S-Graph} minimizes the cost function associated with each of the three layers to estimate the global state of the environment.
Localization is performed by associating the detected planes and rooms, to the walls and rooms of the building extracted from architectural plans respectively. 

\subsubsection{Plane Association} \label{sec:plane_ass_s_graphs}
Building information extraction module of section \ref{sec:bim} creates a two layered scene-graph whose first layers comprises of the walls of the building. As mentioned earlier, walls of the building are represented in the form of planes $\leftidx{^W}{\boldsymbol{\pi}} = [\leftidx{^W}\phi, \leftidx{^W}\theta, \leftidx{^W}d]$. Therefore, observed $x$ vertical walls are matched with all the mapped $x$ vertical walls of the building and similarly for $y$ vertical walls. We check the Mahalanobis distance (Eq.~\ref{eq:plane_error}) between the observed plane and all the walls of the building. If the Mahalanobis distance is less than the threshold, that observation is associated to that particular wall, else a new wall is created. Errors in the observation are thus corrected after plane association.

\subsubsection{Room Association}
Similar to plane association, whenever a room is detected by \textit{S-Graph}, the pose of the a room is matched with the pose of all the rooms of the building, and if the difference between the pose of detected room and a particular room of the building is within a threshold, the observed room is matched to that room. We can safely tune the matching threshold close to the room widths, as rooms do not overlap. This allows us to merge planar walls duplicated due to inaccuracies in plane data association (Section.~\ref{sec:plane_ass_s_graphs}).


%% file: experimental_evaluation.tex
\section{Experimental Evaluation}
\label{experimental_evaluation}
\subsection{Experimental Setup}
We validated our proposed approach in multiple construction environments. In this paper, we performed experiments on single storeys only. We extracted semantic, geometrical and topological information of various building models created in Autodesk Revit software\footnote{\url{https://www.autodesk.com/products/revit/architecture}}. Experiments were performed both in simulated and real environments. The robot platform we used in these experiments was Boston Dynamics \textit{Spot}\footnote{\url{https://www.bostondynamics.com/products/spot}} robot which was equipped with a Velodyne VLP-16 3D LiDAR. \textit{Spot} was tele-operated in both simulated environments and real construction sites to collect data. The entire implementation of the proposed methodology was done in C++, and the experiments were done on an Intel i9 16 core workstation.

\subsubsection{Building Information Extraction}
\label{bim_experiment}
We extracted information from multiple construction models created in Revit software. 3D Revit models were first exported to IFC files. These IFC files were then read by an open source library IFC++\footnote{\url{https://ifcquery.com/}} and BimVision\footnote{\url{https://bimvision.eu/}} software to extract semantic, geometrical and topological information of the building. IFC++ library was used to extract the floors/storeys of the entire building and its elevation value. Then for each storey, the walls and rooms contained within them were extracted.

Once we know which walls and rooms belong to which storey of the building, we extract their geometrical information from BimVision software. BimVision gives the starting location i.e $x, \ y, z$ coordinates of every wall and rooms. This information is stored in a CSV file and then used to generate the topological and metric-semantic layer of the \textit{S-Graph}. This information is published over ROS \cite{ros}, which is then subscribed by the robot.
\begin{table}[!ht]
\centering
\caption{Absolute Trajectory Error (ATE) [m], of our proposed method and two other algorithms \textbf{after convergence} on simulated data. Best results are boldfaced. N.L. means Not Localized}
\begin{tabular}{l | l | c c c c}
\toprule
\textbf{Method} & \textbf{Dataset} &  \textit{R1}  & \textit{R2} & \textit{R3}\\ \midrule
  & BM-1 & \textit{N.L.} & \textbf{0.07} & \textbf{0.06} \\ 
 AMCL & BM-2 & 0.10 & 0.16 & \textit{N.L.} \\
 & BM-3 & 0.07 & 0.09 & - \\ 
\bottomrule
  & BM-1 & 0.20 & 0.21 & 0.22 \\ 
 HDL Localization & BM-2 & 0.16 & 0.14 & \textit{N.L.}\\
 & BM-3 & 0.15 & 0.13 & - \\ 
\bottomrule
  & BM-1 & \textit{N.L.} & \textit{N.L.} & \textit{N.L.}\\ 
 \textit{Without Topological Factor (Ours)} & BM-2 & \textit{N.L.} & \textit{N.L.} & \textit{N.L.} \\ 
 & BM-3 & \textit{N.L.} & \textit{N.L.} & - \\  
\bottomrule
  & BM-1 & \textbf{0.10} & \textit{0.08} & \textit{0.07}\\ 
 \textit{S-Graph Localization (Ours)} & BM-2 & \textbf{0.05} & \textbf{0.07} & \textbf{0.08} \\ 
 & BM-3 & \textbf{0.06} & \textbf{0.07} & - \\  
\bottomrule
\end{tabular}
\label{tab:ate_simulated_data}
\end{table}
\subsubsection{Simulated Experiments} We performed global localization on three  different environments to estimate where the robot started in the building. These environments were created by extracting the meshes from the Revit models of actual architectural models. Mesh files of each storey of building are also extracted from BimVision using IFC file. Theses meshes are then used in simulated experiments. The meshes were imported in Gazebo\footnote{\url{http://gazebosim.org/}} physics simulator to create the replica environments of building models. In the simulated
experiments the odometry was estimated only from LiDAR. The three environments are named as BM-1 (Building Model-1) , BM-2, and BM-3. All of these environments contain multiple rooms in them, and we performed the experiments by placing the robot in different rooms to estimate the robustness of the approach. The legged robot was placed in a random room and then tele-operated to collect data. 

\subsubsection{Real-World Experiments}
We performed experiments on three actual construction sites, the same three environments we created a replica of in the simulation, to estimate if our approach works on real data sets. The legged robot we used in our experiments is equipped with encoders, and we use these encoders to estimate the odometry of the robot.  

\subsubsection{Ablation Study}
We also test our method by removing the novel topological information factor introduced within the particle filter, to evaluate the accuracy of our method. Thus, instead of matching the observations with only the walls of the rooms in which the particle lies, we matched them with all the walls of that storey of the building. 

\label{results_and_discussion}
\begin{table}[!ht]
\centering
\caption{Convergence time (sec) of each algorithm in different environments, starting in different rooms.  Best results are boldfaced. N.L. means Not Localized}
\begin{tabular}{l | l | c c c c}
\toprule
\multicolumn{1}{l}{\textbf{Convergence Time (sec)}}  \\
\toprule
\textbf{Method} & \textbf{Dataset} &  \textit{R1}  & \textit{R2} & \textit{R3}\\ \midrule
  & BM-1 & \textit{N.L.} & 228 & 168 \\ 
 AMCL & BM-2 & 205 & 220 & \textit{N.L.} \\
 & BM-3 & 233 & 197 & - \\ 
\bottomrule
  & BM-1 & 105 & 122 & 70 \\ 
 HDL Localization & BM-2 & 10 & 20 & \textit{N.L.}\\
 & BM-3 & 11 & 23 & - \\ 
\bottomrule
  & BM-1 & \textbf{10} & \textbf{8} & \textbf{12}\\ 
 \textit{S-Graph Localization (Ours)} & BM-2 & \textbf{6} & \textbf{14} & \textbf{9} \\ 
 & BM-3 & \textbf{6} & \textbf{15} & - \\  
\bottomrule
\end{tabular}
\label{tab:time_to_converge}
\vspace{-2mm}
\end{table}
\subsection{Results and Discussion}
\subsubsection{Simulated Experiments}
We compared the performance of out proposed method against two state-of-the-art localization algorithms namely \textit{AMCL} \cite{amcl} and \textit{HDL Localization} \cite{hdl_graph_slam}. Table~\ref{tab:ate_simulated_data} shows ATE of each algorithm in different scenarios. We test each algorithm in three different rooms of each building. \textit{R1}, \textit{R2} and \textit{R3} represent room 1, 2 and 3 respectively. It is to be noted that our algorithm is able to localize and estimate the starting point of the robot in every scenario, whereas \textit{AMCL} fails to localize the robot in multiple scenarios. It should also be noted that without using topological information factor within the paricle filter, our algorithm fails to localize correctly in any environment. 

\subsubsection{Real Experiments}
We tested our algorithm on real data which was collected on actual construction sites of the corresponding buildings. Table~\ref{tab:rmse_real_data} shows the root mean square error between the maps estimated after using different localization algorithms, and, actual mesh of that environment. \textit{AMCL} was unable to localize in the real datasets because the LiDAR measurements in the real environments are quite noisy, and \textit{AMCL} only uses 2D measurements. Since both \textit{HDL Localization} and our approach uses 3D information, they were able to localize successfully. However, our proposed localization method outperformed the other two approaches by considerable margin. Our method performed considerably better than HDL Localization. Moreover, the convergence time of our approach was also better than both \textit{AMCL} and \textit{HDL Localization}.
\begin{table}[ht]
\caption{Point cloud RMSE [m] on the real datasets.Best results are boldfaced, second best are underlined.}
\centering
\begin{tabular}{l | c c c}
\toprule
& \multicolumn{3}{l}{\textbf{Dataset}} \\
\toprule
\textbf{Method} & \textit{BM-1} &  \textit{BM-2} & \textit{BM-3} \\ \midrule
AMCL & \textit{N.L.} & \textit{N.L.} & \textit{N.L.}  \\
HDL Localization & \underline{0.43} & \underline{0.27} & \underline{1.03}  \\
\textit{S-Graph Localization (ours)} & \textbf{0.35} & \textbf{0.25} & \textbf{0.99}  \\
\bottomrule
\end{tabular}
\label{tab:rmse_real_data}
\end{table}

 On real datasets however, we could not test our approach in every room of the construction site due to limited resources. We performed the experiments one room per construction site. but the success of our algorithm in all three construction sites demonstrates that our algorithm can work in any environment irrespective of the room it is started in.
 
\subsubsection{Limitations}
Our proposed localization method shows encouraging results on both simulated as well as real world data when compared against state-of-the-art. However, the current version of our approach does have some limitations. First, we would like to fully automate the building information extraction module, from the loading of the IFC file until the generation of the topological and metric-semantic layer of the \textit{S-Graph}, for the environment. Second, our approach is limited to the robot starting in any given room of the building for the particle filter to converge, although not a strong limitation, we aim to remove this limitation incorporating additional semantic and topological information from the environment. 

%% file: conclusion.tex
\section{Conclusion}
\label{conclusion}
In this paper, we present our novel work for localization of mobile robots on construction environments integrating BIM with scene-graphs. We extract useful geometric, semantic as well as topological information from the readily available BIM plans to integrate it within a three layered hierarchical (\textit{S-Graphs}) \cite{s_graphs}, by generating the metric-semantic and topological layer for a given environment. As the robot navigates in the environment with an on-board 3D LiDAR, our novel implementation of the particle filter utilizing geometric, semantic and topological information rapidly estimates the initial guess of the robot pose in the environment, after which the \textit{S-Graphs} starts creating the robot tracking layer appropriately connecting it with the previously generated metric-semantic and topological layers. We tested our approach in both simulated and real construction environment comparing it with  traditional metric based localization algorithms achieving state-of-the-art results. In future, we plan to improve the information extraction from the BIM as well as localize the robot irrespective of it starting in a room.